\documentclass[10pt,twocolumn,letterpaper]{article}

\usepackage[final]{cvpr} 
\usepackage{hyperref}

\usepackage{multirow} 
\usepackage[table]{xcolor} 
\usepackage{microtype}   
\usepackage{balance}     
\usepackage{stfloats}    

\title{RAM: Recover Any 3D Human Motion in-the-Wild}

\makeatletter
\def\thanks#1{\protected@xdef\@thanks{\@thanks
        \protect\footnotetext{#1}}}
\makeatother

\author{Sen Jia\textsuperscript{\rm 1, *}, Ning Zhu\textsuperscript{\rm 2, *}, Jinqin Zhong\textsuperscript{\rm 2}, Jiale Zhou\textsuperscript{\rm 3}, Huaping Zhang\textsuperscript{\rm 4}, \\ Jenq-Neng Hwang\textsuperscript{\rm 1}, Lei Li\textsuperscript{\rm 4, \dag} \thanks{\(^{*}\) These authors contributed equally to this work.} \thanks{\(^{\dag}\) Corresponding Author. (\href{mailto:lenny.lilei.cs@gmail.com}{\color{black}{\texttt{lenny.lilei.cs@gmail.com}}}) \\
\textsuperscript{\rm 1}University of Washington
\textsuperscript{\rm 2}Anhui University
\textsuperscript{\rm 3}East China University of Science and Technology
\textsuperscript{\rm 4}Beijing Institute of Technology}
}

\begin{document}

\maketitle


\begin{abstract}
RAM incorporates a motion-aware semantic tracker with adaptive Kalman filtering to achieve robust identity association under severe occlusions and dynamic interactions.
A memory-augmented Temporal HMR module further enhances human motion reconstruction by injecting spatio-temporal priors for consistent and smooth motion estimation.
Moreover, a lightweight Predictor module forecasts future poses to maintain reconstruction continuity, while a gated combiner adaptively fuses reconstructed and predicted features to ensure coherence and robustness.
Experiments on in-the-wild multi-person benchmarks such as PoseTrack and 3DPW, demonstrate that RAM substantially outperforms previous state-of-the-art in both Zero-shot tracking stability and 3D accuracy, offering a generalizable paradigm for markerless 3D human motion capture in-the-wild. 
\end{abstract}

\section{Introduction}
Multiple human 3D motion recovery from monocular videos\cite{newell2025comotion,ugrinovic2024multiphys,park2023towards,jiang2024multiply,gordon2022flex,cheng2021graph,zhu2025mesorch,ma2025imdl} aims to robustly track and reconstruct temporally coherent 3D human meshes in real time. Accurate 3D motion data plays a vital role in a wide range of applications, including sports analytics, human–computer interaction, medical rehabilitation, and virtual content creation\cite{li2026multiple,liang2025graphrag, li2025chatmotion}. Traditional motion capture systems, while offering high precision, depend on multi-view camera setups or wearable markers\cite{lee2024mocap,zhang20204d, jia2024adaptive}, which are expensive, intrusive, and difficult to deploy in real-world environments. In contrast, monocular video provides a low-cost and non-invasive alternative that enables markerless 3D motion estimation from easily captured in-the-wild videos\cite{Liu_Kang_2026, Xu_Liu_Mattei_Zheng_2026, liu2026discoveringcontrolinterventionalboundary}. As a result, achieving robust and accurate multi-person 3D motion recovery from monocular videos has become an important and active area of research\cite{ugrinovic2024multiphys, guan2025mcdi,cai2025bayesian,cai2025role}.

\begin{figure}[t]
   \centering
   \includegraphics[width=1\linewidth]{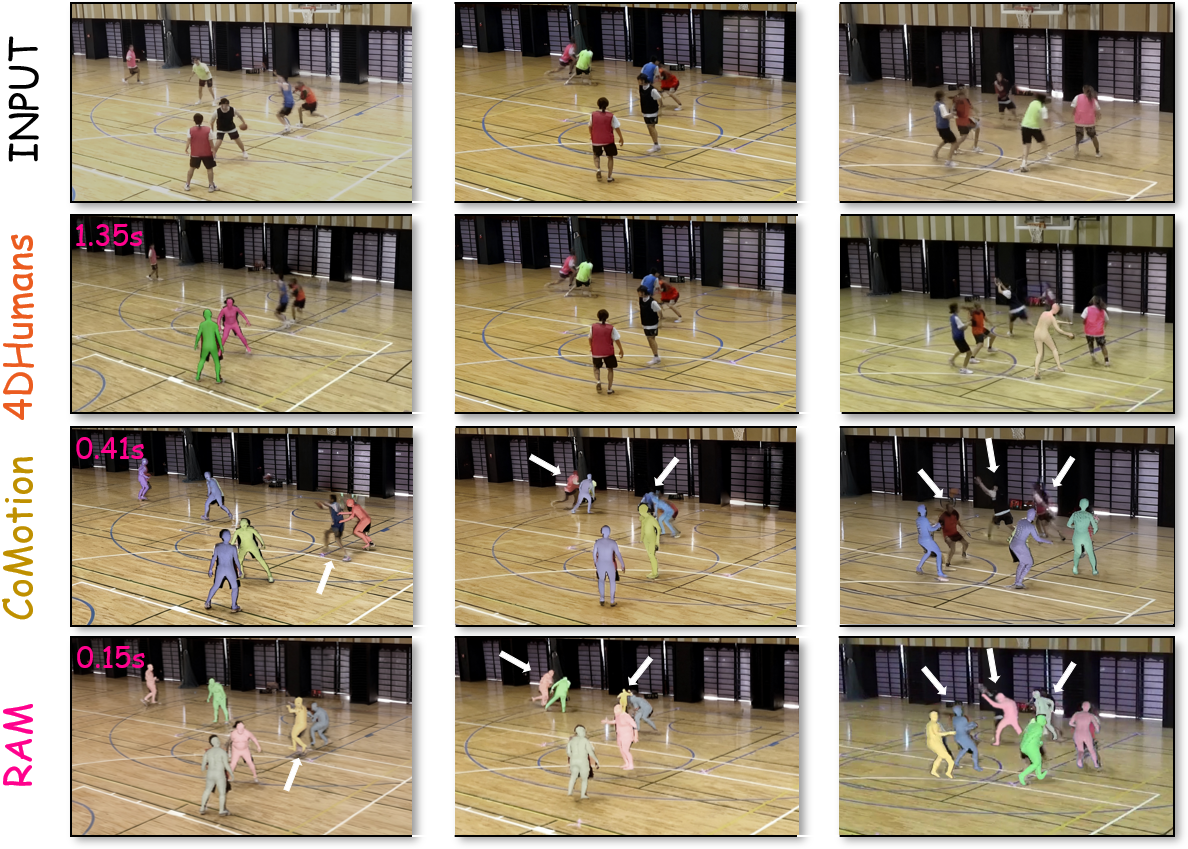}
   \caption{RAM performs online 3D motion reconstruction from monocular video via semantic, motion-aware tracking and occlusion-robust prediction. Unlike prior methods that rely on frame-wise regression and queue-based identity matching, which are often unstable under fast motion and occlusions, RAM maintains consistent identity and accurate reconstruction, while achieving real-time performance that is 2--3$\times$ faster than previous approaches. Arrows highlight key differences. 
}
   \label{fig:motivation}
\end{figure}

Early approaches such as HMR\cite{kanazawa2018HMR}, SPIN\cite{kolotouros2019SPIN}, and PARE\cite{kocabas2021pare} demonstrated single-person 3D mesh reconstruction from static images through end-to-end regression of SMPL\cite{loper2023smpl}. 
Despite these advances, these methods are tailored for single-person scenarios and cannot effectively handle complex, dynamic scenes with multiple interacting individuals. To overcome this limitation, recent research has shifted toward multi-person settings. 4DHuman\cite{goel2023humans} combines HMR2.0\cite{goel2023humans} with PHALP\cite{rajasegaran2022PHALP}-based tracking to support frame-by-frame mesh reconstruction for multiple subjects, while CoMotion\cite{newell2025comotion} jointly optimizes tracking and modeling within an end-to-end framework. Although these methods mark meaningful progress, multi-person motion recovery in-the-wild remains an unsolved challenge\cite{du2026unsupervised, du2026pansharpening, ma2026gor}.
Most existing methods rely heavily on 2D appearance features and the Hungarian algorithm for identity association\cite{bewley2016SORT,wojke2017DeepSORT,li2025human,wang2025reasoningretrievalstudyanswer,li2025frequency}. This strategy is highly sensitive to fast motion, severe occlusion, and viewpoint changes, which often result in identity switches or lost tracks. Once identity continuity is broken, the corresponding 3D motion sequence becomes inconsistent, producing fragmented or duplicated trajectories that degrade reconstruction accuracy.
When targets are partially occluded or undergo fast motion, current methods tend to lose track because they depend solely on visible frame information and lack memory-based motion priors\cite{yu2022towards,doersch2023tapir,yao2025countllm,zhou2024infant,shi2024scaling,shi2025explaining}. This leads to discontinuous or unstable reconstructions once the subject reappears. Furthermore, unstable tracking frequently triggers redundant detection, repeated model initialization, and unnecessary computation, preventing real-time performance and scalability\cite{wang2025selfdestructivelanguagemodel, liu2024graph}.
To address these challenges, we propose the Recover-Anyone Module (\textbf{RAM}), a unified framework for real-time and robust multi-person 3D motion recovery from monocular videos\cite{li2025sepprune,li2026comprehensive}. RAM integrates three complementary components to jointly improve tracking stability, occlusion handling, and temporal coherence.
(1) The SegFollow module introduces\textbf{ motion-aware priors} via adaptive Kalman filtering to ensure stable identity association under fast motion and heavy occlusion.
(2) The Temporal HMR (T-HMR) module enhances 3D reconstruction by injecting \textbf{spatio-temporal cues} from adjacent frames, producing smooth and consistent mesh sequences.
(3) A lightweight Motion \textbf{Predictor} forecasts future poses based on historical motion patterns, maintaining continuity during occluded frames. A \textbf{Combiner} then fuses reconstructed and predicted features to achieve coherent long-term motion recovery.

Extensive evaluations on tracking and recovery benchmarks such as PoseTrack~\cite{andriluka2018posetrack} and 3DPW~\cite{vonMarcard2018}, we demonstrate that RAM achieves state-of-the-art performance on video-based multi-person 3D motion recovery\cite{yang2025you,shi2025medal}. These results underscore RAM's potential to advance human-centric AI, providing a robust foundation for advancing research in motion understanding, social interaction, and multi-agent modeling. Our contributions can be summarized as follows:
\begin{itemize}
    \item We propose \textbf{RAM}, a unified framework that combines motion-aware tracking and temporal mesh recovery to achieve robust, coherent, and occlusion-resilient multi-person 3D motion reconstruction.

    \item RAM achieves state-of-the-art performance in \textbf{Zero-shot tracking stability}, reconstruction accuracy, and inference efficiency across standard benchmarks, providing a solid foundation for diverse downstream tasks\cite{zhang2025invertible}.

    \item RAM is the \textbf{first} method to achieve \textbf{stable, zero-shot multi-person motion recovery} in long, real-world monocular videos with minimal ID switches and without retraining,\textbf{ bridging the sim-to-real gap} and enabling scalable 3D human capture in-the-wild\cite{zhang2025sdarl}.   
\end{itemize}

\section{Related Works}
\paragraph{Tracking and Motion Modeling}
Recent advances in single-object tracking (SOT) have been driven by end-to-end frameworks and stronger appearance modeling~\cite{zhao-etal-2025-tiny,zhang-etal-2025-uora,cao2025pretraining}. Early tracking-by-detection pipelines~\cite{kirillov2023SAM,gu2025mocount,kang2025lp} decouple target localization and temporal association, but their loosely coupled design often leads to error accumulation under long-term or occluded scenarios~\cite{zhou2025edgeTAM,wu2024sc4d,yu2022relationtrack,zhao2025llm}. More recent unified architectures, such as MixFormer~\cite{cui2022mixformer} and SAM2~\cite{ravi2024sam2}, integrate target representation and tracking into a single model and even introduce memory mechanisms to enhance temporal stability~\cite{han2026survey}. While these approaches achieve impressive zero-shot single-target tracking performance, they remain sensitive to heavy occlusions, rapid motion, and large appearance changes, which commonly occur in real-world HOI scenarios. Multi-Object tracking methods follow a tracking-by-detection paradigm, combining object detections with identity association across frames\cite{11078508,zhao2024large}. Classical approaches~\cite{rajasegaran2022PHALP} leverage Kalman filtering~\cite{kalman1960KF}, Hungarian matching~\cite{kuhn1955hungarian}, or handcrafted motion cues, while recent works such as Tracktor~\cite{bergmann2019Tracktor}, MotionTrack~\cite{qin2023motiontrack}, and MambaTrack~\cite{xiao2024mambatrack} adopt learned appearance, motion fusion for improved trajectory consistency~\cite{butepage2017deep,lu2024towards,chen2025llmmentalhealth}. However, MOT systems generally rely on dataset-specific detectors, or domain-specific finetuning. As a result, they remain far from robust in in-the-wild settings with crowding, fast motions, and domain shifts.

\paragraph{Human Motion Reconstruction}
Human motion estimation has progressed from single-frame, single-person reconstruction to multi-person, video-based human recovery. Early single-frame approaches~\cite{moreno20173d,bogo2016keep,du2025forensichub} focus on estimating 3D body pose or mesh from cropped human regions\cite{ma2025finegrained, ma2025msdetr, 10.1145/3726302.3730070}, following either parametric pipelines such as HMR~\cite{kanazawa2018HMR} and its successors (e.g., 4D Humans), which regress SMPL parameters~\cite{loper2023smpl,jiang2024back}, or non-parametric methods that directly predict mesh vertices\cite{ji2026retrieval, Ji2025, ji2025mrag, Zhao_Min_Wu_Li_Sun_Cai_Wang_Chen_Penn_2026}. While effective for isolated subjects, these approaches struggle in crowded scenes. Recent works therefore adopt transformer-based architectures to reason across multiple human instances within an image, improving robustness in multi-person settings~\cite{dong2024panocontext,hiller2024perceiving,li2022maskfpan,soro2024d2nwg}. Building on single-frame reconstruction, video-based human motion estimation extends the challenge to temporal association and long-term identity consistency~\cite{hu2021self,khoiee2025multi,shen2023global}. Tracking-by-detection remains the dominant paradigm, where 2D detections are linked across frames using appearance cues~\cite{chaabane2021deft,li2022time3d,li2025wav2sem,jiang2025unihpr}. Systems such as PHALP~\cite{rajasegaran2022PHALP}, 4D Humans~\cite{goel2023humans} and CoMotion~\cite{newell2025comotion} further incorporate 3D features to enhance association robustness under occlusions, while attention-based matching strategies have been explored to improve identity stability~\cite{jiang2025unihpr,li2025aublendshape}. However, these methods still rely on dataset-specific tuning and motion matching, limiting their generalization and robustness in in-the-wild scenarios.

\section{Method}
\label{sec:method}

\begin{figure*}[t]
   \centering
   \includegraphics[width=0.75\linewidth]{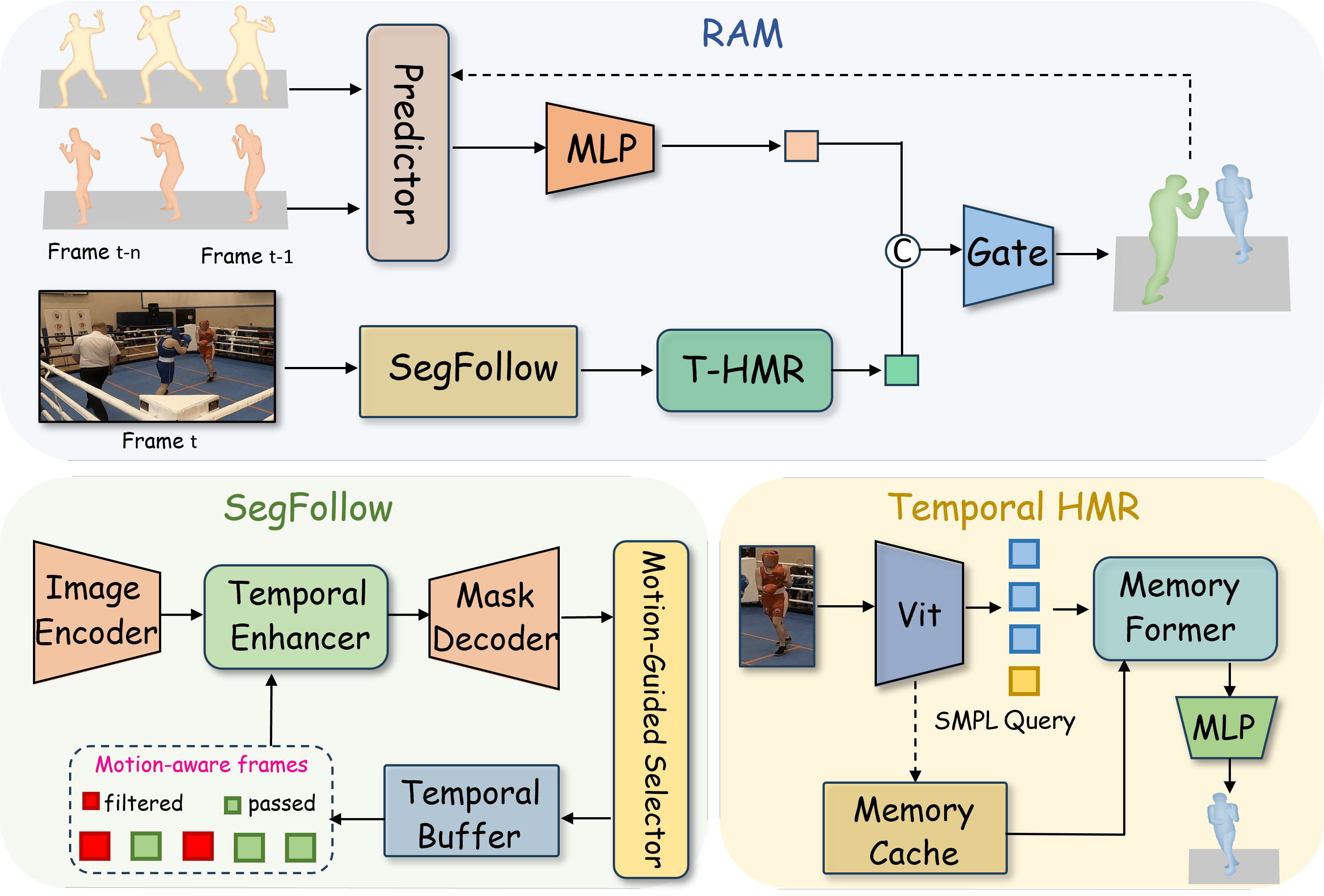}
   \caption{Overview of RAM. The framework integrates four components: SegFollow for motion-guided temporal tracking, Temporal HMR for memory-based 3D reconstruction, a Predictor for motion forecasting under occlusion, and a gated Combiner for robust recovery.}
   \label{fig:overview}
\end{figure*}


RAM is a unified framework for real-time and accurate multi-person 3D human motion reconstruction from monocular videos. As illustrated in Figure~\ref{fig:overview}, RAM consists of four key components:
\begin{enumerate}
    \item \textbf{SegFollow} performs motion-aware semantic tracking to maintain stable identity associations across frames, even under occlusion and rapid motion.
    \item \textbf{T-HMR} reconstructs 3D human meshes from the tracked instances by incorporating temporal context via memory-augmented attention, enabling coherent and robust mesh estimation.
    
    \item \textbf{Predictor} models motion dynamics from past reconstructions and forecasts future pose states, offering strong priors when current observations are unreliable.
    
    \item \textbf{Combiner} fuses predictions and reconstructions through a learnable gating mechanism, producing stable SMPL outputs with improved temporal consistency under uncertainty.
\end{enumerate}

\subsection{SegFollow Module}
\label{subsec:segfollow}
The SegFollow module builds upon SAM2 to enable robust identity tracking across frames. While SAM2 offers strong segmentation performance, its naive FIFO-based memory update lacks temporal reliability modeling, often leading to identity switches and noise accumulation. SegFollow addresses this by introducing two components: a \textbf{motion-guided selector}, which combines Kalman filtering and motion-aware scoring for reliable mask association; and a \textbf{temporal buffer}, which updates memory with confidence-weighted smoothing to preserve temporal continuity. Together, they augment SAM2 with explicit motion reasoning and stable long-term tracking.

\subsubsection{Motion-Guided Selector}
\label{subsubsec:motion_selector}
While long-term temporal aggregation is critical for consistent tracking, unreliable associations can corrupt memory with noisy features. To mitigate this, we introduce a motion-guided selector that fuses appearance similarity from SAM2 with motion-aware prior, enabling a more robust and explicit motion reasoning to obtain reliable temporal associations. 
For a detected instance at time step $k$, we represent its observation as a bounding box 
$\mathbf{z}_k = [x_k, y_k, w_k, h_k]^T$
and model the underlying motion state as:
\[
\mathbf{x}_k = [x_k, y_k, w_k, h_k, \dot{x}_k, \dot{y}_k, \dot{w}_k, \dot{h}_k]^T.
\]
To estimate the motion state of the object, we apply a Kalman Filter (KF), which models temporal dynamics under Gaussian noise and linear assumptions. Given the previous state, the KF predicts a bounding box $\mathbf{H}\hat{\mathbf{x}}_k^{-}$ for the next frame and estimates its uncertainty. The filter is conditionally updated only when the observation is sufficiently reliable, allowing robust state propagation under partial or noisy observations.

\paragraph{Motion–Aware Selection.}
For each candidate segmentation mask $M_i$ from SAM2, its bounding box $\mathbf{z}_{k,i}$ is compared to the KF prediction via an IoU-based motion-consistency score:
\[
s_{\text{kf}}(M_i) = \mathrm{IoU}\!\left(\mathbf{H}\hat{\mathbf{x}}_k^{-},\, \mathbf{z}_{k,i}\right).
\]
We combine this score with SAM2's mask affinity $s_{\text{mask}}(M_i)$ via a gated sum:
\[
s_{\text{fused}}(M_i) = \alpha\, s_{\text{mask}}(M_i) + (1-\alpha)\, s_{\text{kf}}(M_i),
\]
The mask with the highest $s_{\text{fused}}$ is selected. This fusion improves association robustness in challenging scenes, enabling SegFollow to maintain stable tracking even under fast motion or occlusion.

\paragraph{Confidence-Gated Update.}
To avoid updating the motion state with noisy observations, we adopt a confidence-gated update strategy. Let $s_{\text{obj}}(M^\star)$ denote the confidence score for the chosen mask, generated by the mask decoder. If the confidence exceeds a threshold, we increment a counter $C_k$ that tracks consecutive reliable associations. 
The KF posterior is updated only when sufficient evidence has accumulated ($C_k \ge \tau_{kf}$); otherwise, the previous state is retained:
\[
\hat{\mathbf{x}}_k^{+} =
\begin{cases}
\hat{\mathbf{x}}_k^{-} + \mathbf{K}_k\!\left(\mathbf{z}_k - \mathbf{H}\hat{\mathbf{x}}_k^{-}\right), & C_k \ge \tau_{kf}, \\[2pt]
\hat{\mathbf{x}}_{k-1}^{+}, & \text{otherwise}.
\end{cases}
\]
This gated update prevents unreliable detections, such as during occlusion or rapid motion, from corrupting the motion state, yielding more stable identity tracking compared to SAM2’s original FIFO-based memory update.

\subsubsection{Temporal Buffer}
\label{subsubsec:temporal_buffer}
To further enhance temporal stability, we design a temporal buffer that adaptively updates the memory bank $\mathcal{B}_t$ with smooth weighting. Unlike the FIFO strategy in SAM2, our update mechanism formulates memory update as an exponential moving average, adaptively modulated by motion confidence. At each time step $t$, the memory embedding $\mathcal{B}_{t-1}$ and the current key feature $K_t$ are combined as:
\[
\mathcal{B}_t = \gamma_t \mathcal{B}_{t-1} + (1 - \gamma_t) K_t,
\]
where $\gamma_t = 1 - \min(s_{\text{kf}}(M^*), \, \tau_\gamma)$ is an adaptive decay factor modulated by the Kalman-based motion consistency score. The decay factor $\gamma_t$ adaptively balances current and historical cues: reliable motion prompts stronger updates from present features, while uncertain motion preserves past memory to maintain temporal consistency. This update scheme selectively incorporates high-confidence frames into the memory, guiding attention with temporally reliable context. 
By incorporating the motion-guided selector and temporal buffer, \textbf{SegFollow enables zero-shot, occlusion-robust object tracking}. This lightweight yet effective design provides a solid foundation for stable and accurate 3D reconstruction downstream.

\subsection{T-HMR}
\label{subsubsec:t_hmr}
Given the temporally tracked instances from SegFollow, T-HMR aims to reconstruct 3D human meshes by regressing SMPL parameters from each frame. As using only current-frame features leads to temporal inconsistency and weak robustness under occlusion, T-HMR introduces two core components: \textbf{Memory Cache} for selecting informative and reliable temporal features, and the \textbf{MemFormer} for injecting these priors into the reconstruction process. Details are provided below.

\subsubsection{Memory Cache}
\label{subsubsec:memory_cache}
To retrieve useful temporal priors, we design a Memory Cache that adaptively selects the top-$k$ relevant frame features from a temporal window centered at the current frame. Specifically, we collect the features encoded by ViT from $L$ adjacent frames, and stack them into a memory feature $F_{\text{mem}} \in \mathbb{R}^{L \times N \times d}$, where $N$ is the number of spatial tokens and $d$ is the feature dimension. 
We first pool the current frame representations and memory feature across spatial dimensions into global representations \( \bar{F}_t \) and \( \bar{F}_{\text{mem}} \in \mathbb{R}^{L \times d}\), respectively. Then, we adopt a \textbf{dual-branch scoring mechanism} to rank the importance of each frame in $F_{\text{mem}}$. A unified attention-based scoring function is defined as:
\[
\mathcal{A}(F_q, F_k) = 
\text{softmax}_N \left[
\frac{(F_q W_q)(F_k W_k)^\top}{\sqrt{d}}
\right] \in \mathbb{R}^{N \times N},
\]
where $F_q, F_k \in \mathbb{R}^{N \times d}$ and $W_q, W_k \in \mathbb{R}^{d \times d}$.
For the \textbf{first branch}, we compute the attention scores between the current frame and memory frames to measure relevance and dependencies. For the \textbf{second branch}, we evaluate the internal consistency of memory frames through self-attention. The overall importance score is then obtained by combining both branches. Formally:
\[
s = \mathcal{A}(\bar{F}_t, \bar{F}_{\text{mem}}) + \text{mean}\left(\mathcal{A}(\bar{F}_{\text{mem}}, \bar{F}_{\text{mem}})\right),
\]
Finally, the top-$k$ frames with highest overall importance scores are selected for the memory bank. By modeling both cross-frame relevance and intra-frame consistency, Memory Cache distills high-quality temporal cues while discarding redundant frame features. This facilitates reliable reconstruction under occlusion and motion blur, while ensuring efficiency for MemFormer reasoning.

\subsubsection{MemFormer}
Given the selected memory features from the Memory Cache and the current frame feature, MemFormer aims to integrate temporal priors for SMPL regression. The overall architecture is composed of $N$ stacked blocks, each following the design described below.
We first concatenate a learnable SMPL token with the current frame feature and apply self-attention to model intra-frame interactions. The resulting representation serves as the query to perform temporal cross-attention, where the memory features are pooled along the spatial dimension to provide motion-consistent keys and values. The output is then passed to a second cross-attention block, which uses memory pooled along the temporal dimension to inject spatially aligned semantics from recent frames.
Finally, the updated SMPL token is extracted and decoded via an MLP head to regress the SMPL pose and shape parameters for the current frame. This design allows T-HMR to inject fine-grained temporal priors into the current-frame reconstruction process, enhancing the model's robustness to occlusion and smoothness. 

\subsection{Predictor}
\label{subsec:predictor}
To enhance robustness under occlusion, the Predictor forecasts future motion based on recent reconstruction history. Let $\mathbf{S}_t$ denote the reconstructed motion state at time $t$, we maintain a FIFO queue $\mathcal{Q}_t = \{\mathbf{S}_{t-T+1}, \dots, \mathbf{S}_t\}$. This sequence is fed into a stack of $L$ Transformer blocks to capture motion dynamics and predict a latent representation $\hat{Z}{t+1}$ for the next frame. The output latent $\hat{Z}_{t+1}$ is then passed to the Combiner module as a motion-conditioned prior, supporting stable reconstruction when current-frame cues are unreliable. The motion queue is updated online during inference, enabling real-time adaptation to dynamic motion patterns.

\subsection{Combiner}
\label{subsec:combiner}
The Combiner integrates the motion prior from the Predictor and the current reconstruction feature from T-HMR to produce a stable next-step motion. It fuses the representations via a learnable gate and regresses SMPL parameters in a single head.
Given the T-HMR feature $Z_{t+1}^{\mathrm{h}}$ of the current frame and the predicted latent motion prior $\hat{Z}_{t+1}$ from the Predictor, the Combiner predicts a gating vector via an MLP layer:
\[
g_{t+1} = \sigma\!\left(\mathrm{MLP}_g\!\left([Z_{t+1}^{\mathrm{h}}, \hat{Z}_{t+1}]\right)\right) \in [0,1]^d,
\]
where $\sigma(\cdot)$ is the sigmoid function. The fused feature is then computed using a weighted interpolation:
\[
Z_{t+1}^{\mathrm{c}} = (1 - g_{t+1}) \odot Z_{t+1}^{\mathrm{h}} + g_{t+1} \odot \hat{Z}_{t+1},
\]
where $\odot$ denotes element-wise multiplication. This design encourages reliance on T-HMR features under confident observations, while shifting toward the predicted prior under occlusion or uncertainty, enabling stable and consistent motion recovery. Finally, a regression head maps $Z_{t+1}^{\mathrm{c}}$ to SMPL parameters for next-frame reconstruction. 

\subsection{Training Objectives}
We adopt a three-stage learning strategy to train RAM.
\textbf{Stage1}. 
Following 4D-Humans, we pretrain T-HMR’s image encoder and pose regression module on large batches of single images. The SMPL supervision employs multiple objectives: L1 loss on 2D joint projections and pelvis-relative 3D joint positions, L2 loss on joint rotations, L1 loss on shape parameters (betas, for fully annotated samples), binary cross-entropy loss for prediction confidence (with unmatched outputs as negatives), and a keypoint heatmap loss for spatial consistency.

\textbf{Stage2}. 
We train the temporal predictor with a scheduled sampling strategy exclusively on 8-frame 3D pose sequences and synthetic 3D pose sequences generated by panning/zooming single-image pose data from InstaVariety. The 3D and 2D pose points of the first frame are matched to high-quality ground-truth annotations, and the predictor is unrolled over time to perform temporal prediction of pose states up to the n-th frame. Each timestep is supervised using a multi-term loss that jointly optimizes 3D pose point accuracy and 2D pose point projection consistency (consistent with the SMPL-based loss framework of Stage 1).

\textbf{Stage3}. 
With the Predictor and RAM components frozen, we fine-tune the full framework to learn dynamic fusion under occlusion. We simulate occlusion by randomly masking 60\% of human body regions in the input images, encouraging the Combiner to rely more on the predicted motion prior when visual cues are incomplete. This strategy promotes robust recovery by guiding the model to adaptively balance reconstruction and prediction. Supervision follows the Stage-1 SMPL loss.

\begin{table*}[t]
    \centering
    \caption{Comparison on PoseTrack18 and PoseTrack21: while 4DHumans and CoMotion are trained on these datasets, RAM is evaluated zero-shot without retraining, and achieves the best results across all metrics.}
    \label{tab:posetrack_results}
    \resizebox{\linewidth}{!}{
        \begin{tabular}{lcccccccccc}
            \toprule
            \multirow{2}{*}{Method} & \multicolumn{3}{c}{PoseTrack18} & \multicolumn{7}{c}{PoseTrack21} \\
            & HOTA$\uparrow$ & IDs$\downarrow$ & MOTA$\uparrow$ & MOTA$\uparrow$ & IDF1$\uparrow$ & IDP$\uparrow$ & IDR$\uparrow$ & FP$\downarrow$ & FN$\downarrow$ & FPS$\uparrow$ \\
            \midrule
            4DHumans \cite{goel2023humans} & 57.8 & 382 & 61.4 & -- & -- & -- & -- & -- & -- & - \\
            4DHumans (reproduced)$^*$ & 58.0 & 349 & \textbf{61.8} & 56.7 & 70.9 & 87.1 & 59.7 & 7817 & 50652 & 0.51 \\
            CoMotion~\cite{newell2025comotion} \textit{strict} & 58.2 & 232 & 59.9 & 61.8 & 74.0 & 89.1 & 63.3 & 6086 & 45664 & - \\
            CoMotion~\cite{newell2025comotion} & 54.9 & 344 & 51.3 & 71.4 & 79.5 & 87.1 & 73.0 & 8115 & 30394 & 5.68 \\
            \midrule
            \rowcolor{green!8}RAM (\textbf{Zero-shot}) & \textbf{66.4} & \textbf{15} & 57.7 & \textbf{74.4} & \textbf{85.9} & \textbf{93.8} & \textbf{79.2} & \textbf{5864} & \textbf{24044} & \textbf{10.32} \\
            \bottomrule
        \end{tabular}
    }
\end{table*}

\begin{table*}[t]
    \centering
    \caption{Comparison on two challenging real-world sports scenarios featuring frequent occlusions, and fast motion. We evaluate zero-shot generalization of prior methods and RAM, and additionally report an ablation using only SAM2-based tracking.}
    \label{tab:zeroshot}
    \resizebox{0.85\linewidth}{!}{
        \begin{tabular}{lcccccccccc}
            \toprule
            \multirow{2}{*}{Method} & \multicolumn{3}{c}{TrackID3x3 (Indoor)} & \multicolumn{3}{c}{TrackID3x3 (Outdoor)}\\
            & TI-HOTA$\uparrow$ & TI-DetA$\uparrow$ & TI-AssA$\uparrow$ & TI-HOTA$\uparrow$ & TI-DetA$\uparrow$ & TI-AssA$\uparrow$ \\
            \midrule
            4DHumans~\cite{goel2023humans} & 5.17 & 4.89 & 5.47 & 2.66 & 2.19 & 3.22  \\
            CoMotion~\cite{newell2025comotion} & 42.20 & 32.93 & 54.07 & 30.87 & 23.06 & 41.33  \\
            \midrule
            RAM (SAM2) & 54.23 & 48.39 & 60.78 & 38.60 & 36.92 & 40.37  \\
            \rowcolor{green!8}RAM & \textbf{75.07} & \textbf{62.87} & \textbf{89.66} & \textbf{66.68} & \textbf{51.39} & \textbf{86.63} \\
            \bottomrule
        \end{tabular}
    }
\end{table*}
\balance

\section{Experment}
\label{sec:experiment}

To evaluate the effectiveness of RAM, we conducted comprehensive experiments on multiple datasets, assessing both its tracking and estimation capabilities. Our evaluation includes standard benchmarks as well as challenging scenarios characterized by rapid motion, frequent occlusions, and the presence of multiple interacting subjects such as in basketball and boxing scenes. 

\subsection{Implementation Details}
\subsubsection{Evaluation Datasets}
We evaluate RAM across multiple benchmarks covering 2D/3D pose estimation and multi-person tracking. For tracking and 2D keypoint estimation, we use the PoseTrack~\cite{andriluka2018posetrack} and COCO dataset~\cite{lin2014microsoft}, which feature frequent occlusions, complex motions and diverse poses. In addition, TrackID-3x3~\cite{yamada2025trackid3x3} and Olympic Boxing dataset further challenge robustness in real-world sports with dense interactions and fast motion, serving as rigorous benchmarks for evaluating generalization and robustness. For 3D reconstruction, we adopt 3DPW~\cite{vonMarcard2018}, which offers accurate mesh and joint annotations in unconstrained outdoor settings.

\subsubsection{Evaluation Metrics}
Following the evaluation protocol of previous works~\cite{newell2025comotion}, we report standard metrics including Multiple Object Tracking Accuracy (MOTA), Identity F1 Score (IDF1), number of ID switches (IDs), ID precision and recall (IDP, IDR), and Hybrid Object Tracking Accuracy (HOTA). We also include error statistics such as false positives and false negatives for comprehensive analysis. Among these, MOTA primarily evaluates the completeness of detection, whereas IDF1, IDs, IDP, and IDR quantify how well the tracker maintains identity consistency over time. For TrackID-3x3, we additionally report TI-HOTA, TI-DetA, and TI-AssA~\cite{yamada2025trackid3x3}, which evaluate ID-consistent tracking by disentangling detection and association accuracy. For 2D pose estimation on COCO~\cite{lin2014microsoft} and PoseTrack~\cite{andriluka2018posetrack}, we use the Percentage of Correct Keypoints (PCK), which measures the proportion of keypoints predicted within a normalized distance threshold. For 3D pose evaluation on 3DPW~\cite{vonMarcard2018}, we report the Mean Per Joint Position Error (MPJPE) and its Procrustes-aligned variant (PA-MPJPE), which account for absolute and rigid-aligned joint accuracy, respectively.

\balance
\subsection{Results}
\label{subsec:tracking}
\subsubsection{Tracking Results}
\label{subsubsec:tracking_results}

\begin{table*}[t]
\centering
\caption{\textbf{Pose estimation.} Normalized PCK accuracy on projected 2D keypoints at varying thresholds on the COCO and PoseTrack datasets, alongside MPJPE of 3D keypoints on the 3DPW dataset. We highlight that our model performs similarly when provided the full image as input rather than an oracle-resized crop around a target person. See text for analysis.
}
\label{tab:2d_pck}
\caption{Your caption here (e.g., Quantitative comparison on 2D and 3D pose estimation benchmarks).}
\centering
\resizebox{\linewidth}{!}{
\begin{tabular}{@{}ll cc@{\hspace{6mm}} cc@{\hspace{6mm}} cc@{}}
\toprule
& Method & \multicolumn{2}{c}{COCO} & \multicolumn{2}{c}{PoseTrack} & \multicolumn{2}{c}{3DPW} \\
& & PCKn@0.05$\uparrow$ & PCKn@0.1$\uparrow$ & PCKn@0.05$\uparrow$ & PCKn@0.1$\uparrow$ & MPJPE$\downarrow$ & PA-MPJPE$\downarrow$ \\
\midrule
\multirow{9}{*}{\rotatebox[origin=c]{90}{2D/3D HPE}} 
& PyMAF~\citep{zhang2021pymaf}          & 0.68 & 0.86 & 0.77 & 0.92 & 92.8 & 58.9 \\
& CLIFF~\citep{li2022cliff}             & 0.64 & 0.88 & 0.75 & 0.92 & 69.0 & 43.0 \\
& PARE~\citep{kocabas2021pare}          & 0.72 & 0.91 & 0.79 & 0.93 & 82.0 & 50.9 \\
& PyMAF-X~\citep{zhang2023pymaf}        & 0.79 & 0.93 & 0.85 & 0.95 & 78.0 & 47.1 \\
& HMR 2.0a~\citep{goel2023humans}       & 0.79 & 0.95 & 0.86 & 0.97 & 70.0 & 44.5 \\
& HMR 2.0b~\citep{goel2023humans}       & 0.86 & 0.96 & 0.90 & \textbf{0.98} & 81.3 & 54.3 \\
& Comotion~\citep{newell2025comotion}   & 0.79 & 0.92 & 0.88 & 0.96 & 60.0 & 37.3 \\
\midrule
\rowcolor{green!8}
& \textbf{RAM (Ours)}                   & \textbf{0.89} & \textbf{0.97} & \textbf{0.93} & \textbf{0.98} & \textbf{53.0} & \textbf{34.1} \\
\bottomrule
\end{tabular}
}
\end{table*}

We compare against state-of-the-art 3D recovery methods including 4DHumans and CoMotion. While other multi-object tracking methods exist~\cite{lv2024diffmot, yang2024hybrid}, they are not tailored for 3D recovery and typically require tuning for domain transfer. For completeness, we include comparisons in the supplementary, where our SegFollow still achieves superior zero-shot tracking ability.

\paragraph{Evaluation on PoseTrack}
We evaluate RAM on the PoseTrack benchmarks to assess its tracking performance under challenging scenarios involving occlusion, fast motion, and frequent identity interactions. As shown in Table~\ref{tab:posetrack_results}, RAM consistently outperforms prior state-of-the-art methods such as 4DHumans and CoMotion across key dimensions, including identity consistency, multi-person detection coverage, and robustness. Due to their reliance on 3D trajectory matching, both 4DHumans and CoMotion exhibit poor performance under occlusion or viewpoint shifts, often resulting in identity switches and fragmented tracks.

In contrast, RAM attains a HOTA of 66.4 with merely \textbf{15 ID switches} on PoseTrack18, marking an \textbf{order-of-magnitude} gain in identity stability over prior methods; the slight drop in MOTA stems largely from stricter association behavior rather than tracking failures. On PoseTrack21, RAM sets a new benchmark with 74.4 MOTA, and further achieves \textbf{+6.4} IDF1 and\textbf{ +82\%} FPS improvements over CoMotion, establishing RAM as the \textbf{new state-of-the-art for real-time, stable human motion tracking} in-the-wild. These gains stem from the SegFollow module, which integrates motion-aware priors via a Kalman-based selector for robust mask association, and employs a temporal buffer that focuses historically reliable frames, ensuring stable identity propagation even under occlusion and motion blur. 

These results highlight RAM’s effectiveness in enabling robust and consistent multi-person tracking in wild, real-world video scenarios. Notably, \textbf{RAM achieves this impressive performance in a zero-shot setting, without retraining on PoseTrack}, highlighting its remarkable generalization to in-the-wild tracking scenarios.

\paragraph{Zero-Shot Evaluation on Challenging Real-World Videos}
We evaluate RAM's zero-shot generalization on TrackID-3x3~\cite{yamada2025trackid3x3}, a challenging basketball competition dataset from real-world featuring frequent occlusions and fast motion. Compared to PoseTrack’s short clips ($\sim$3s), TrackID-3x3 includes longer videos ($\sim$10s indoor, $\sim$40s outdoor), making the outdoor setting more challenging due to video length and in-the-wild complexity.

As shown in Table~\ref{tab:zeroshot}, RAM significantly outperforms prior methods, achieving \textbf{+78\%} (indoor) and \textbf{+116\%} (outdoor) higher TI-HOTA than CoMotion. These gains highlight RAM’s robustness in real-world scenarios with frequent occlusion and multi-person interaction. In contrast, 4DHumans and CoMotion underperform due to their reliance on domain-specific trajectory matching.

Our \textbf{ablation with SAM2}-based tracking alone results in notable performance drops, particularly in outdoor scenarios, despite using a strong pre-trained segmenter. While SAM2 provides semantic initialization, it lacks motion priors and temporal modeling, leading to unstable tracking under occlusions and fast motion. Its TI-HOTA on outdoor TrackID-3x3 only improves over CoMotion by \textbf{+7.7}, indicating limited generalization. In contrast, full RAM yields a \textbf{+35.8} gain, demonstrating that our \textbf{motion-aware designs are key} to achieving robust, real-world tracking and cannot be replaced by strong segmentation alone.

Together, these results establish RAM as the first framework to robustly generalize to long, in-the-wild multi-human videos under zero-shot settings.

\subsubsection{Estimation Results}
\label{subsubsec:estimation_results}

\paragraph{Evaluation on PoseTrack and COCO}
We evaluate RAM on the PoseTrack and COCO datasets to assess its reconstruction accuracy in real-world complex scenarios. These 2D datasets encompass diverse human motions and challenging real-world conditions such as crowd interactions and occlusion, providing a rigorous benchmark for evaluating model robustness and generalization. As shown in Table~\ref{tab:2d_pck}, RAM achieves state-of-the-art PCK accuracy across both datasets, outperforming prior works including PARE, CLIFF, and HMR 2.0. Notably, RAM achieves \textbf{0.93} PCK@0.05 on PoseTrack and \textbf{0.89} on COCO, indicating strong localization precision even under dense, multi-person settings. Compared to CoMotion, which already integrates temporal cues, RAM still yields consistent gains, highlighting its temporal stability and occlusion robustness.

These gains primarily derive from T-HMR’s temporal priors and the Predictor’s motion-conditioned inference, which together enable robust estimation under occlusion. The Combiner further refines this by adaptively balancing predicted and observed cues, ensuring consistent recovery even when joints are fully invisible. These results highlight RAM’s effectiveness in delivering stable and accurate pose estimation across complex, real-world conditions.

\begin{figure*}[t]
   \centering
   \includegraphics[width=0.95\linewidth]{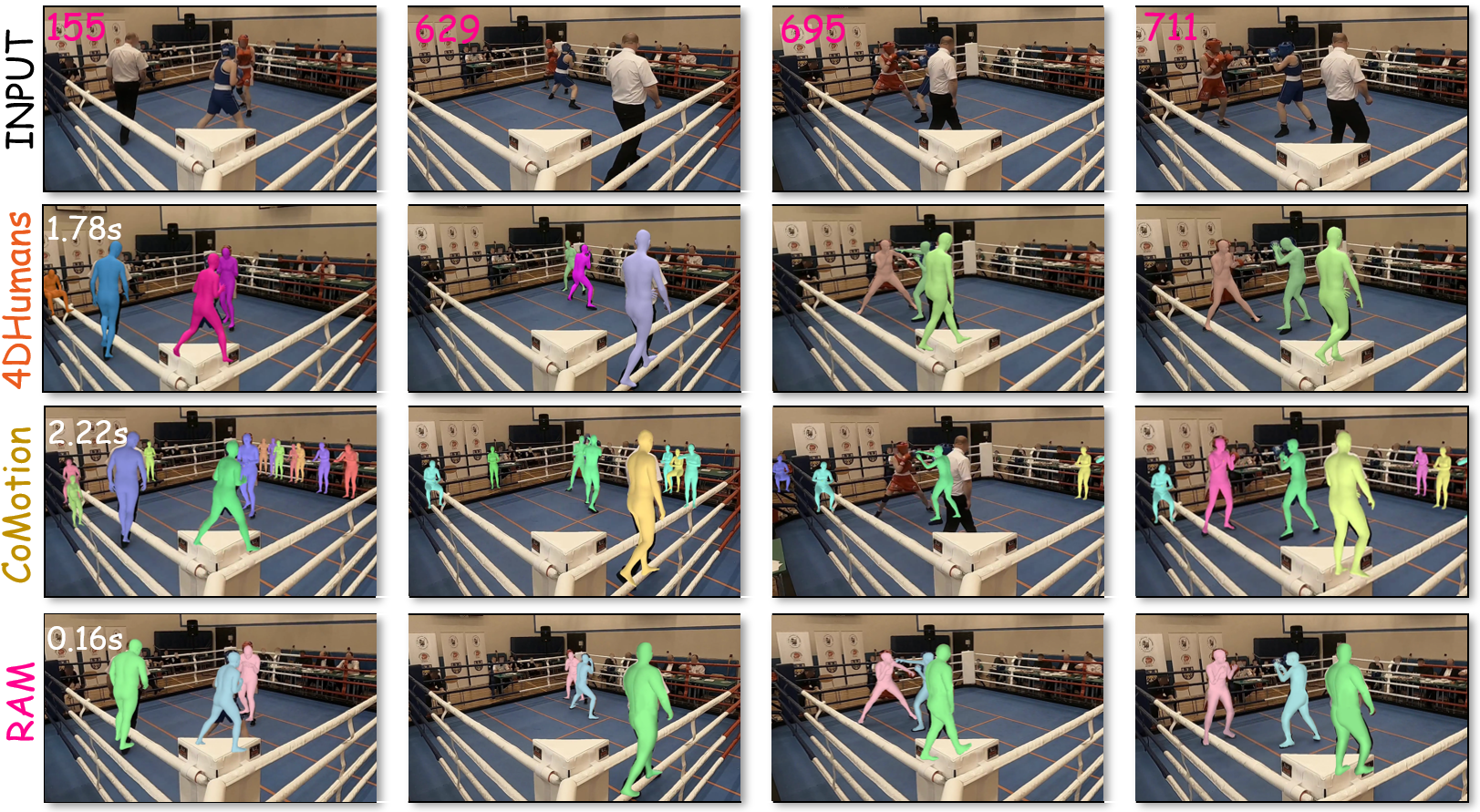}
   \caption{Qualitative comparison on the Olympics Boxing dataset. Both 4DHumans and CoMotion suffer from identity switches and tracking failures under fast motion and heavy occlusion, resulting in fragmented 3D reconstructions and repeated identity reinitialization. This not only degrades reconstruction quality but also leads to high inference overhead. In contrast, RAM robustly tracks boxers and the referee throughout the sequence with consistent identity association and real-time, accurate 3D motion recovery.}
   \label{fig:qualitative}
\end{figure*}

\paragraph{Evaluation on 3DPW}
We further evaluate RAM on the 3DPW dataset to assess its 3D recovery accuracy. Following the CoMotion protocol, we report standard 3D metrics including MPJPE and PA-MPJPE, as shown in Table~\ref{tab:2d_pck}. RAM achieves the lowest reconstruction error among all methods, with \textbf{53.0} MPJPE and \textbf{34.1} PA-MPJPE, outperforming CoMotion and recent single-person models such as HMR 2.0. 
The improvement primarily stems from T-HMR’s temporal priors, which strengthen frame-to-frame consistency. In addition, the synergy between current-frame reconstruction and predictive motion priors further enhances robustness. These results demonstrate that RAM is capable of performing accurate 3D human mesh reconstruction directly from monocular videos, enabling robust and coherent multi-person motion recovery in-the-wild.

\subsubsection{Qualitative Results}
We evaluate RAM on two real-world sports datasets with fast motion, occlusion, and complex multi-person interaction. As shown in Fig.~\ref{fig:overview} and Fig.~\ref{fig:qualitative}, existing methods such as 4DHumans and CoMotion struggle to maintain identity consistency (Person colors are used to denote their tracking IDs) and recover accurate 3D motion. CoMotion often fails under occlusion or motion blur, while 4DHumans exhibits frequent identity switches and unstable queue management, resulting in poor reconstruction and high latency. These limitations prevent existing methods from achieving robust zero-shot multi-human motion recovery in real-world settings.

In contrast, RAM achieves real-time, robust tracking and accurate 3D recovery throughout. Its motion-aware tracking and occlusion-resilient prediction allow it to sustain identity and reconstruction quality even in long and dynamic videos such as boxing sequences.

These results demonstrate RAM's strong \textbf{zero-shot generalization} and its \textbf{practical applicability} for 3D motion recovery in unconstrained real-world settings.

\section{Conclusion}
\label{sec:conclusion}

We propose RAM, a unified framework for real-time and robust multi-person 3D motion reconstruction from monocular videos. By integrating semantic tracking with motion-aware modeling, RAM effectively mitigates ID switches and tracking loss under occlusion and viewpoint changes, enabling stable identity association and reducing redundant computation. Leveraging temporal priors and motion prediction, RAM achieves smooth and accurate reconstruction even under severe occlusions. Extensive experiments demonstrate that RAM consistently outperforms prior works, especially in complex real-world sports scenes, achieving substantial gains in tracking stability, reconstruction quality, and computational efficiency. RAM provides a solid foundation for future research in human-centric motion understanding and will be extended to human-object recovery.

\section{Acknowledgements}
This work was supported in part by the National Key Research and Development Program of China under Grant 2024YFC3308101; in part by the National Natural Science Foundation of China under Grant 62401005; and in part by the Natural Science Foundation of Anhui Higher Education Institutions of China under Grant 2023AH050069.

{\small
\bibliographystyle{ieeenat_fullname}
\bibliography{reference}
}

\end{document}